\ifx\mIncludeGuard\undfined
\input mac/m-impl
\def\subend{\end}
\else
\def\subend{}
\fi

\centerline{\bf Evaluation and Optimization of Leave-one-out Cross-validation for the Lasso}
\smallskip
\centerline{\sl Ryan Burn}
\bigskip

\begingroup
\advance \leftskip by 15pt
\advance \rightskip by 15pt
\ninepoint
  \noindent {\bf Abstract.} I develop an algorithm to produce the piecewise quadratic that computes 
  leave-one-out cross-validation for the lasso as a function of its hyperparameter. 
  The algorithm can be used to find exact hyperparameters
  that optimize leave-one-out cross-validation either globally or locally, and its
  practicality is demonstrated on real-world data sets. I also show how the algorithm can be
  modified to compute approximate leave-one-out cross-validation, making it suitable for larger
  data sets.
  \medskip
\endgroup

\ifx\mIncludeGuard\undfined
\input mac/m-impl
\def\subend{\end}
\else
\def\subend{}
\fi

\newsec{Introduction} Given observations $(\x\idx1, y\idx1)$, $\ldots$, $(\x\idx{n}, y\idx{n})$, 
$\x\idx{i} \in {\bbb R}^p$, $y\idx{i} \in {\bbb R}$, and a shrinkage parameter $t\ge0$, the 
{\it least absolute shrinkage and selection operator} (lasso) produces a solution 
to the optimization problem
$$
  \argmin_{\alpha, \rg} \left\{
    \sum_{i=1}^n \left(y\idx{i} - \alpha - \x\idxp{i}\rg\right)^2\right\}
    \,\hbox{such that}\, \sum_{j=1}^p |\beta_j| \le t.
$$
For $t$ large enough, the lasso is equivalent to ordinary least squares (OLS); for smaller $t$,
the lasso shrinks regression coefficients and can set some coefficients to zero. 
If $t$ is chosen well, the lasso can both improve on the prediction accuracy of OLS and 
reveal more important regressors.
The key to success with the lasso is estimating how out-of-sample performance is related to
$t$ and selecting a good value for the shrinkage parameter. 

Arguably leave-one-out cross-validation (LO) is one of the best methods available to estimate prediction 
accuracy. It has lower bias than other forms of $v$-fold cross-validation [\citecv, \citelo]; 
for many data sets, it is uniquely defined as a function of $t$;
and contrary to what's often claimed, its variance may be no worse than that of $v$-fold cross-validation with 
$v<n$ 
[\citecvsurvey, \citeloci, \citecv].
My aim with this paper is to
develop an algorithm, {\it leave-one-out least-angle regression} (LO-LARS), to efficiently produce 
the piecewise quadratic function that computes
LO exactly for any value of $t$. From there, selecting a good value of the shrinkage parameter 
is an easy step. We might, 
for example, choose the value of $t$ that minimizes LO globally, or we might choose a smaller
value of $t$ with near optimum LO if we want to achieve greater sparsity. 
With the piecewise quadratic components of LO, it's straightforward to go from
a strategy for selecting $t$ to an exact value.
\smallskip
The lasso appears to have been first used by Fadil Santosa and William
Symes in 1986 in an algorithm to deconvolve band-limited reflection
seismograms [\citegeo].
In reflection seismology, the earth is probed with a seismic wavelet. As the wavelet
propagates, it reflects when it passes through geologic interfaces separating layers.
The resulting seismogram can be viewed as the convolution of the known source wavelet with an 
unknown sparse reflector series with spikes corresponding to the geologic interfaces
[\citeseismology, \S3.3].
Using a linear approximation of the wave equation and the lasso, the convolution can be inverted to estimate the reflector
series and reveal subsurface properties. Use of the lasso with its $\ell_1$ restriction on
coefficients makes the 
inversion robust to noise and forces a sparse solution. In their algorithm, Santosa and Symes left the 
choice of the lasso shrinkage parameter to the user, suggesting that it could be selected 
interactively, writing,
\beginquote
The choice of the parameter to weigh the data is left to the user of the algorithm, and
hence the method is suited for interactive data interpretations.
\endquote

A decade after Santosa and Symes' paper, Robert Tibshirani proposed the lasso as a general-purpose
tool for the estimation of linear models [\citelasso]. Tibshirani highlighted the ability of the lasso to perform both 
regularization and variable selection as advantages over other methods such as ridge regression
or subset selection. Tibshirani additionally proposed three different
methods to select the lasso shrinkage parameter:
$$\vbox{\noindent
\item{i)} Estimate prediction error using $v$-fold cross-validation and select $t$ so
as to minimize prediction error across a grid of evenly space values from $0$ to
$\sum_{j=1}^p |\hat{\beta}_j^{\rm{OLS}}|$ where ${\rghat^{\rm{OLS}}}$ represents
the least squares estimate.
\item {ii)} Same as (i), but use an approximation of generalized cross-validation to estimate 
prediction error.
\item {iii)} Select $t$ to minimize Stein's unbiased estimate of risk.
}$$
Benchmarking the three methods with simulated data,  Tibshirani reported,
\beginquote
Estimation of the lasso parameter by generalized cross-validation seems to perform the best, a
trend that we find is consistent through all of our examples.
\endquote
Note that generalized cross-validation is a leave-one-out cross-validation
on a rotated version of the data set [\citegcv].

In an earlier paper, Prabir Burman also studied the effectiveness of different forms of 
cross-validation
 for estimating prediction error [\citecv].
Although Burman did not explicitly look at the lasso, he developed a general
theory for the asymptotic performance of cross-validation that applies to
the lasso as a special case. Burman showed that if the number of regressors $p$ is fixed and
$n\to\infty$, then, under suitable conditions,
$$
{\bbb E} \bigg[
\left(
\hbox{$v$-fold cross-}\atop\hbox{validation error}
\right) -
\left(
\hbox{prediction error}
\right)
\bigg]
= {c_0\over{n(v - 1)}} + {\cal O}({1\over n^2})
$$
where the parameter $c_0$ depends on the von Mises derivative of the error estimator.
It's clear from the formula that bias decreases as the number of folds increases
and that cross-validation obtains its lowest asymptotic bias with $n$ folds where the bias is
${\cal O}({1\over{n^2}})$. Burman also derived a formula for the asymptotic variance of LO,
showing
$$
{\bbb V} \bigg[
\hbox{LO}
-
\left(
\hbox{prediction error}
\right)
\bigg]
= {\cal O}({1\over n}).
$$

Because of the low bias of LO and the ability of the lasso to select more important variables, there's 
been a lot of recent interest in using the lasso with LO in high-dimensional settings where
$p$ is either greater than or close to $n$. Kamiar Rad, Haolin Zou, Arnab Auddy, and Arian Maleki 
proved that versions of
Burman's bounds can be extended to certain generalized linear models and nonsmooth regularizers
when the ratio $p/n$ remains 
constant as $n\to\infty$ 
[\citelop, \citelo].
Additionally using the matrix inversion lemma, Rad, Wang, and Zhou developed an efficient approximation
to LO for regularized generalized linear models [\citealo, \citealop].
\smallskip
The LO-LARS algorithm integrates the techniques used in efficient LO approximation with
a version of the LARS algorithm [\citelars, \citelarslike].
Like LARS, LO-LARS works by incrementally building
lasso solution paths as a function of $t$, starting from $t=0$; unlike LARS, though, it builds solution
paths for every leave-one-out subset of the data set, which are used to produce leave-one-out
errors. 
By using the matrix inversion lemma, LO-LARS is able to eliminate much of the overhead that
would be incurred if the lasso subproblems were to be solved independently;
and by integrating LO computation into a lasso solver,
 LO-LARS is able to account for activation changes in 
the subproblems and compute LO exactly. While exact computation of LO is practical for many data
sets, it can still be prohibitively expensive for some larger data sets. In such cases, LO-LARS
can be adapted to compute the piecewise function for approximate leave-one-out cross-validation [\citealo,
\citealop] instead where individual activation changes are no longer tracked, making the algorithm significantly
less expensive.
\subend

\medskip
\ifx\mIncludeGuard\undfined
\input mac/m-impl
\def\subend{\end}
\else
\def\subend{}
\fi

\newsec{Overview} Before getting into details, I think it will be helpful to look at how LO-LARS works
on a few examples. Consider the diabetes data set from 
the paper by Efron, Hasti, Johnstone, and Tibshirani that
introduces the LARS algorithm [\citelars]:%
\displaytable{.1in}{table/diabetes-preview}{.}{%
Preview of the diabetes data set from [\citelars]. The data set
measures the progression of diabetes in 442 patients over the course of a year. The 
10 regressors are measured at the start of the year.
}
The data set quantitatively measures the progression of diabetes in 442 patients after a year.
At the start of the year, 10 baseline variables are measured. Efron et al.\ explain the modeling 
objective as,
\beginquote
Two hopes were evident here, that the model would produce accurate baseline predictions of
response for future patients and that the form of the model would suggest which covariates were
important factors in disease progression.
\endquote
LO-LARS computes the following LO function for the data set:
\displayfigure{-.5in}{figure/diabetes-1/plot}{.}{%
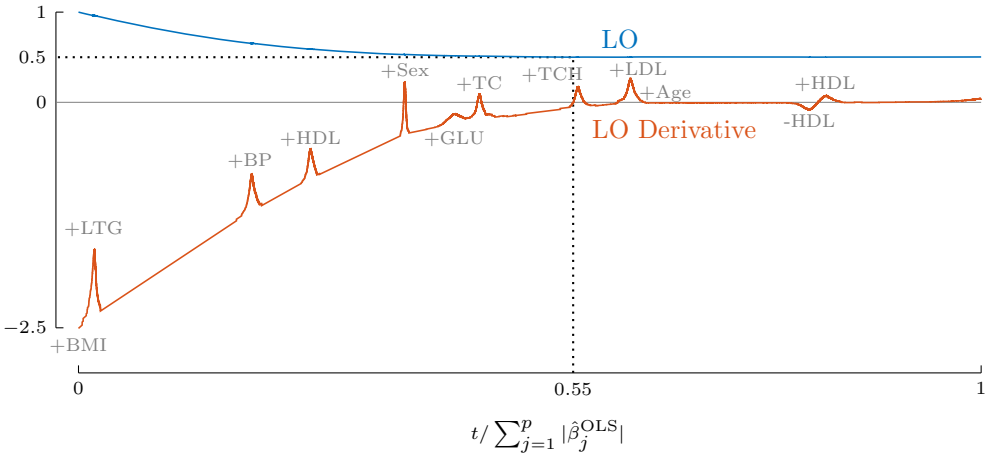
Plot of LO and its derivative for the diabetes data set from [\citelars].
Mean squared error is normalized to have a maximum value of 1 and $t$ is
normalized by the $\ell_1$ norm of the OLS coefficients. The points of
activation and deactivation for each of the ten regressors are shown. We can
see that each spike in the LO derivative is associated with a change in
activation. At its minimum, ${\rm LO} \approx 0.5$.
}
LO is a continuous piecewise quadratic; and its derivative is, therefore, a
piecewise linear function with jump discontinuities. Each point of discontinuity in the derivative 
corresponds to a change in activation of
a leave-one-out subproblem. From the segments, it's easy to compute all the local minimums,
\displaytable{0.1in}{table/diabetes-optimums}{.}{%
Local optimums of LO for the diabetes data set. Also shown are the nonzero regressors
at each optimum.
}
LO-LARS can also produce individual leave-one-out errors as a byproduct,
\displayfigure{-1in}{figure/diabetes-err/plot}{.}{%
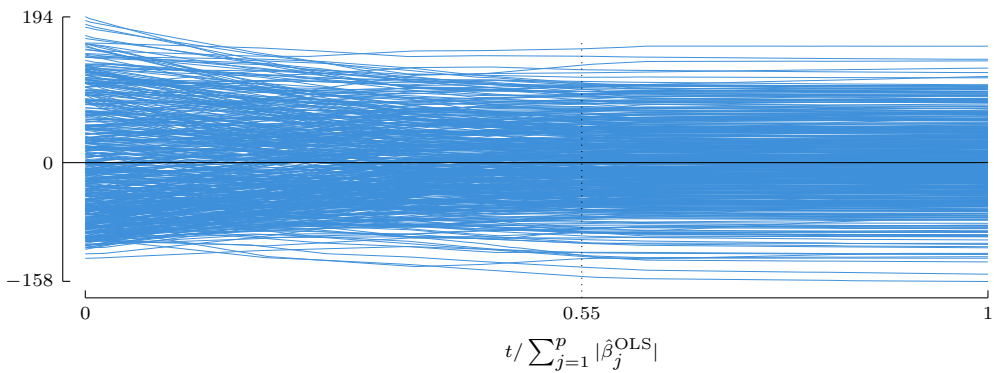
Individual leave-one-out errors, actual progression minus predicted progression,
as a function of the shrinkage parameter
for each of the 442 patients
in the diabetes data set.
}

For many data sets, computing LO fully is entirely feasible. However for
high-dimensional data sets where $p \gg n$, computing LO for larger values of $t$ can become
expensive as individual solution paths won't overlap as much in their active
regressors. Fortunately, it's easy to modify LO-LARS to detect local minimums as it
incrementally expands the domain of LO and to bail out of the algorithm if LO increases past
a given threshold of the best optimum found so far. With this modification, many higher dimensional
data sets also become tractable.
For example, this data set measures riboflavin levels and the expression of $4088$ genes
in cultures of the bacteria {\it Bacillus subtilis\/}:%
\displaytable{0.1in}{table/riboflavin-preview}{.}{%
Riboflavin data set initially from [\citeriboflavinfirst] and made available by
[\citeriboflavin]. The data set measures log riboflavin production rate
of the bacteria Bacillus subtilis
and log transcription levels for 4088 genes. 
}
Riboflavin is known to regulate the transcription of certain genes, and a lasso model can help
to identify which genes are influenced by riboflavin.
If I instruct LO-LARS to exit early if LO exceeds 1\% of the best optimum found so
far, I can compute the LO function below up to the early stopping point
in approximately half a second. 
By comparison, computing the whole LO function takes slightly over 3 seconds.
\displayfigure{-.5in}{figure/riboflavin/plot}{}{%
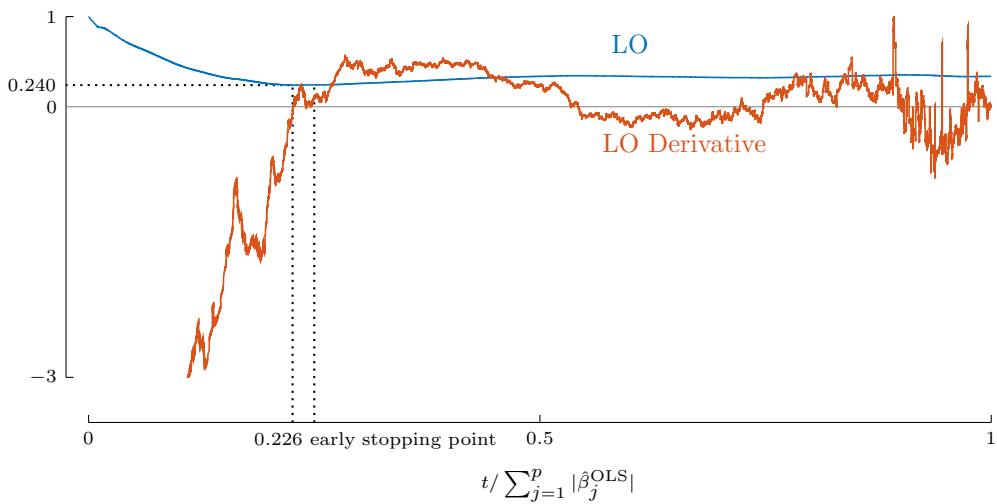
Plot of LO and its derivative for the riboflavin data set from [\citeriboflavin].
Mean squared error is normalized to have a maximum value of 1 and $t$ is
normalized by the $\ell_1$ norm of the OLS coefficients. At its optimum,
$t / \sum_{j=1}^p |\rghat^{OLS}| \approx 0.226$, $LO \approx 0.24$, and the regressors
for 20 genes are active. If we 
instruct LO-LARS to stop early with a 1\% threshold, it still finds the global optimum
and aborts at $t / \sum_{j=1}^p |\rghat^{OLS}| \approx 0.25$.
}
As we can see from the graph, the LO derivative gets quite choppy for larger values of $t$. The
choppiness
corresponds to divergence in the sets of active regressors for the leave-one-out solution paths,
which makes LO more expense to compute.

For much larger data sets, it can make more sense to use approximate leave-one-out
cross-validation (ALO). ALO is significantly less expensive to compute as it doesn't 
require LO-LARS to track activation changes for each individual data entry and ALO becomes more
accurate as the size of the data set increases.
\displayfigure{-.5in}{figure/alo-comp/plot}{}{%
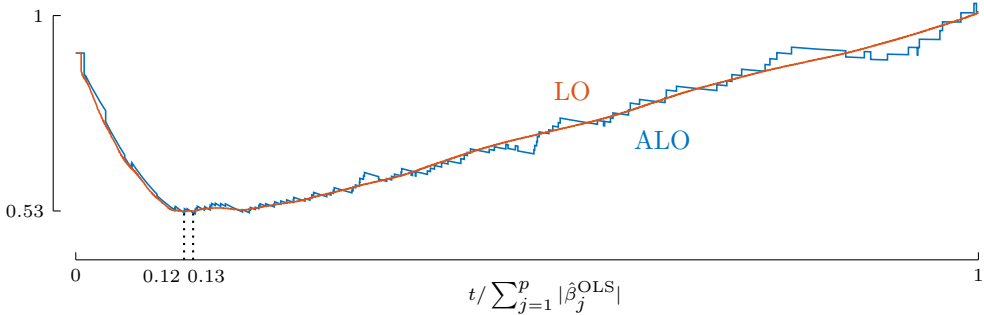
Comparison of ALO and LO for an example data set of dimensions $n=200$ and $p=100$. The
design matrix is drawn from a normal distribution with a Toeplitz correlation matrix of 
parameter $0.5$; the first 10 true regressors are set to 1 and the remaining true regressors
are set to zero; and the errors are normally distributed with variance set so that the
signal-to-noise ratio is 1.
LO is minimized at 
$t / \sum_{j=1}^p |\rghat^{OLS}| \approx 0.12$ 
and ALO is 
minimized at $t / \sum_{j=1}^p |\rghat^{OLS}| \approx 0.13$.
}
The figure above compares ALO and LO for an example data set. We can see that
the two curves track each other closely with their optimum values differing by about 8 percent.
LO-LARS took around $0.2$ seconds to compute the LO function and around $0.01$ seconds to compute the
ALO function.
% ALO takes ~.013sec, LO takes ~.19sec

The table below shows benchmarking results for a variety of real-world data sets of varying dimensions.
\displaytable{0.1in}{table/benchmark}{}{
Benchmarks of a variety of real world data sets. Shown are the dimensions of the data set,
the number of active regressors at the optimum, the shrinkage of the optimum, LO at optimum
(expressed relative to LO with maximal shrinkage), and the average elapse over 5 runs.
{\tt Q} indicates that the original regressors were expanded with quadratic terms and
{\tt E} indicates that LO-LARS was stopped early with a 1\% threshold. Details
are given in Appendix A.
}
We can see that LO-LARS is fast enough to be practical for many data sets. To get \hfil\break
a better understanding of what determines its performance, we can look at how long it\hfil\break
takes to fit certain simulated data sets.

Consider the linear model
$$\displaylines{
y = \x' \rgtrue + \varepsilon,\cr
\x\sim N(\nullv_p, \Sigma)\quad\hbox{with}\quad\Sigma_{ij} = 0.8^{|i-j|},\quad\hbox{and}\cr
\varepsilon\sim N(\nullv_n, \sigma^2 {\bf I})\quad\hbox{with}\quad\sigma^2 = \rgtrue' \Sigma \rgtrue.
}$$
Suppose that the first $k$ regressors of $\rgtrue$ are drawn from a standard Laplace distribution
and the remaining $p-k$ regressors are set to $0$. The table below shows how long it takes me to
run LO-LARS using LO and ALO with an early exit threshold of 1\% for all combinations of
$ n, \,p\in\left\{\,800,\, 1600\,\right\}$ and $k\in\left\{\,10,\, 50\,\right\}$.
\displaytable{-0.7in}{table/bench-sim2}{}{
Benchmarks for fitting different simulated data sets with LO-LARS and an early exit threshold of 1\%. 
Each entry is averaged over 10 runs.
Shown are the dimensions of the simulated linear model,
the number of nonzero true regressors, $k$, the average number of nonzero regressors of the LO
optimum, the relative prediction error computed as 
$(\rghat - \rgtrue)'\Sigma(\rghat-\rgtrue)/\rgtrue'\Sigma\rgtrue$,
and the average elapse time to run LO-LARS for both LO and ALO.
}
\subend

\medskip
\ifx\mIncludeGuard\undfined
\input mac/m-impl
\def\subend{\end}
\else
\def\subend{}
\fi

\newsec{Solution Paths} Given $\y \in \Re^n$, $\X\in\Re^{n\times p}$, 
and $\lambda \ge 0$, consider
$$
  {1\over2}
     \left\|\y - \X \rg\right\|^2 + \lambda \sum_{j=p'+1}^p \left| \beta_j\right|.
     \filleqno{obj}
$$
If $\rghat$ minimizes (\eqobj), then $\rghat$ solves a lasso problem with the first $p'$ regressors unrestricted
and the $\ell_1$ norm of the remaining regressors bounded above by $\sum_{j=p'+1}^p |\hat{\beta}_j|$.
By allowing for unrestricted regressors, we can incorporate the intercept term into cross-validation 
estimates, and we can solve certain generalized lasso problems [\citegenerallasso].
We'll call a function $\rghat(\lambda):[0, \infty) \to \Re^p$ a {\it solution path} if
$\rghat(\lambda)$ is continuous and
$$
\rghat(\lambda) \in \argmin_{\rg}
\left\{
  {1\over2}
     \left\|\y - \X \rg\right\|^2 + \lambda \sum_{j=p'+1}^p \left| \beta_j\right|
\right\}\qquad\hbox{for all $\lambda\ge0$.}
$$
Consider, for example, the data set
$$\matrix{
x_1 & \phantom{-}0.09 & -0.88 & -1.77 & -0.10 & 1.00\cr
x_2 & \phantom{-}0.01 & \phantom{-}0.91 & -1.04 & \phantom{-}0.81 & 0.27\cr
y & -0.09 & -1.57 & -1.47 & -1.08 & 1.49\cr
}\;.$$
The piecewise linear solution path is specified by the nodes
$$\matrix{
\lambda & 
\phantom{-}0\phantom{.000} & 1.412 & 5.573 & \infty\cr
x_1 & \phantom{-}1.321 & 0.845 & 0\phantom{.000}& 0\cr
x_2 & -0.757 & 0\phantom{.000} & 0\phantom{.000}& 0\cr
}\;.\filleqno{solnex}$$
Solution paths need not be unique. For instance, if we duplicate a regressor,
$$\matrix{
x_1 & \phantom{-}0.09 & -0.88 & -1.77 & -0.10 & 1.00\cr
x_2 & \phantom{-}0.01 & \phantom{-}0.91 & -1.04 & \phantom{-}0.81 & 0.27\cr
x_3 & \phantom{-}0.01 & \phantom{-}0.91 & -1.04 & \phantom{-}0.81 & 0.27\cr
y & -0.09 & -1.57 & -1.47 & -1.08 & 1.49\cr
}\;,$$
then clearly (\eqsolnex) is still a solution path, but we can easily form infinitely many
additional solution paths from weighted averages of $x_2$ and $x_3$.
\smallskip
We can use convex optimization theory to derive some basic properties of solution paths. Define
  $\L(\rg) \eqdef {1\over2}\left\|\y - \X \rg\right\|^2$,$\,$
  $r(\rg) \eqdef \sum_{j=p'+1}^p |\beta_j|$, $\,$
  $f(\rg; \lambda) \eqdef \L(\rg) + \lambda \, r(\rg)$, and
$$
  \activer(\rg) \eqdef \fndef{\hsize=3.5in
  The ordered vector that corresponds to the set
  $$\left\{\,1, \ldots, p'\,\right\}\cup
  \left\{\,j\hbox{ such that $\beta_j\ne0$}\,\right\}.$$}
$$
The following theorem is adopted from Ryan Tibshirani [\citeunique].
\smallskip
\noindent {\bf Theorem P.}
\begingroup
\it  If $\lambda\ge0$, then 
\medskip
\item{1.} There exists a $\rghat$ such that
$\rghat$ minimizes $f(\,\cdot\,; \lambda)$ and 
$$\rank\X_{\activer(\rghat)} = \#\activer(\rghat)$$
where \# provides the length of a vector.
\smallskip
\item{2.} If $\rghat_1$ and $\rghat_2$ both minimize $f(\,\cdot\,; \lambda)$, then
$\L(\rghat_1) = \L(\rghat_2)$ and $r(\rghat_1) = r(\rghat_2)$ if $\lambda > 0$.
\smallskip
\item{3.} The vector $\rghat$ minimizes $f(\,\cdot\,; \lambda)$ if and only if
$
  \X'\left(\y - \X \rghat\right) = \lambda \,\gammab
$
for some $\gammab$ where
$$
  \gamma_j \in \cases{
    \{\,0\,\}, & \hbox{\rm if $j\le p'$;}\cr\noalign{\vskip1pt}
    \{\,\sign \beta_j\,\}, & \hbox{\rm if $j > p'$ and $\beta_j \ne 0$;}\cr\noalign{\vskip1pt}
    \left[-1, \,1\right], & \hbox{\rm otherwise.}
  }
$$
\endgroup
\smallskip
% Proof P1
\noindent {\it Proof of P1.}  If $\lambda=0$, 
let $\A$ denote a set of regressors such that $\X_\A$ has maximal rank. Set $\rghat$ so that
$\rghat_{-\A} = \nullv$ and $\rghat_\A = \left(\X_\A'\X_\A\right)^{-1} \X_\A' \y$.
Then $\rghat$ minimizes $f(\,\cdot\,;\lambda)$ and the nonzero entries identify regressors of full rank.

The next part of the proof follows Saharon Rosset [\citesparse]. Suppose $\lambda>0$. Assume first that the unrestricted regressors are of full rank.
Clearly, $f(\rg;\lambda)\to\infty$ as $\| \rg\|\to\infty$. Choose $R$
so that
$f(\rg; \lambda) > f(\nullv;\lambda)$ if $\|\rg\|>R$.
Since 
$$
B_R(\nullv) = \big\{\,\rg\hbox{ such that $\|\rg\|\le R$}\,\big\}
$$
is compact and $f$ is continuous, $f$ attains a minimum, $\rghat$, in $B_R(\nullv)$.
Suppose that
$$
  \rank\X_\A < \#(\A)\qquad\hbox{for }
  \A = \activer(\rghat).
$$
Then we can pick $\u \neq \nullv$ such that
$\u_{-\A} = \nullv$ and $\X_{\A}\u_{\A} = \nullv$.
Put
$
  g(t) = f(\rghat + t\u; \lambda)
$. Then $g$ is differentiable in a neighborhood of $0$ and $g'(0) = 0$; otherwise we would have
$f(\rghat + \delta \u; \lambda) < f(\rghat; \lambda)$ for some $\delta$. Put
$$
\hat{t} = \argmin_t |t|\;\hbox{
such that $\left(\rghat + t \u\right)_j = 0$ for some $j\in\A$ and $j>p'$}
$$
and form a new minimum with fewer active regressors by setting $\rghat \leftarrow \rghat + \hat{t} \u$. If 
the nonzero coefficients of $\rghat$ 
still do not identify regressors of full rank, repeat the process until the full rank condition is satisfied.

Finally, assume the first $p'$ regressors are not of full rank. Let $\cal F$ be a subset of
$\{1, \ldots, p'\}$ identifying regressors of full rank such that  $\rank \X_{\cal F}$ is
maximized. Put ${\cal F} \leftarrow {\cal F} \cup \{p'+1, \ldots, p\}$. Set $\rghat_{-\cal F}=\nullv$
and set $\rghat_{\cal F}$ to be a full rank minimizer of the lasso problem for $(\X_{\cal F}, \y)$.
\qed
\smallskip
% Proof P2
\noindent {\it Proof of P2.} Suppose that $\rghat_1$ and $\rghat_2$ are two minimums of 
$f_\lambda$ and $ \L(\rghat_1) \neq \L(\rghat_2)$. Put
$$
  g(t) = \L(\rghat_1 + t(\rghat_2 - \rghat_1)).
$$
Then
$$
  g(t) = g(0) + g'(0) t + {1\over2} g''(0) t^2\quad \hbox{and}\quad g''(0) > 0.
$$
Now,
$$
\eqalign{%
{1\over2}\left(g(0) + g(1)\right) &= g(0) + {1\over2}g'(0) + {1\over4}g''(0)\cr
             &> g(0) + {1\over2}g'(0) + {1\over8}g''(0) = g({1\over2}).
}
$$
Thus,
$$
\eqalign{
  f(\half \rghat_1 + \half\rghat_2; \lambda)
      &= \L(\half \rghat_1 + \half \rghat_2) + \lambda\, r(\half\rghat_1 + \half \rghat_2)\cr
      &= g(\half) + \lambda\, r(\half\rghat_1 + \half \rghat_2)\cr
      &<  \half \L(\rghat_1) + \half \L(\rghat_2) + \lambda\, r(\half\rghat_1 + \half \rghat_2).
}
$$
By the triangle inequality, $r(\rg_1 + \rg_2) \le r(\rg_1) + r(\rg_2)$; hence,
$$
\eqalign{
  f(\half \rghat_1 + \half\rghat_2; \lambda)
      &<  \half 
      \left(\L(\rghat_1) + \lambda\, r(\rghat_1)\right) + 
      \half \left(\L(\rghat_2) + 
      \lambda\,r(\rghat_2)\right)\cr
  &= \half f(\rghat_1; \lambda) + \half f(\rghat_2; \lambda).
}
$$
But since $\rghat_1$ and $\rghat_2$ are both minimums, $f(\rghat_1; \lambda) = f(\rghat_2; \lambda)$;
and $\half\left(\rghat_1 + \rghat_2\right)$ must therefore be a minimum smaller than the others---a contradiction.
We conclude that $\L(\rghat_1) = \L(\rghat_2)$ and likewise $\lambda\,r(\rghat_1) = \lambda\,r(\rghat_2)$. 
\qed
\smallskip
% Proof P3
To prove P3, we make use of subgradients [\citeconvex, \S23]. If $f:\Re^n\to\Re$ is a convex function, then a subgradient 
of $f$ at a point $\x_0$ is a vector $\u\in\Re^n$ such that
$$
f(\x) - f(\x_0) \ge \u' (\x - \x_0)\quad\hbox{for all}\quad \x\in\Re^n.
$$
The set of all subgradients of $f$ at $\x_0$, denoted by $\partial f(\x_0)$, is called a 
subdifferential. Clearly, $\x_0$ is a minimum of $f$ if and only if
$\nullv \in \partial f(\x_0)$.
\smallskip
\noindent {\it Proof of P3.} We have $|x| \ge c x$ for all $x$ if and only if
$c \in [-1, 1]$. Thus, the subdifferential of $r$ at $\rg$ is given by
$$
\gammab \in \partial r(\rg) \quad\hbox{if and only } \gamma_j\in \cases{
    \{\,0\,\}, & \hbox{\rm if $j\le p'$};\cr\noalign{\vskip1pt}
    \{\,\sign \beta_j\,\}, & \hbox{\rm if $j > p'$ and $\beta_j \ne 0$;}\cr\noalign{\vskip1pt}
    \left[-1, \,1\right], & \hbox{\rm otherwise;}
}\qquad\hbox{for all $j$}.
$$
The subdifferential of $\L$ is given by its gradient,
$$
\partial \L(\rg) = \left\{\,-\X'\left(\y - \X \rg\right)\,\right\}.
$$
So, $\nullv \in \partial f(\rghat; \lambda)$ is equivalent to
$$
\X' \left(\y - \X \rghat\right) = \lambda\,\gammab\quad\hbox{for some $\gammab\in\partial r(\rghat)$,}
$$
which establishes the result.\qed

\subend

\medskip
\ifx\mIncludeGuard\undfined
\input mac/m-impl
\def\subend{\end}
\else
\def\subend{}
\fi

\newsec{LARS Algorithm} In this section, we develop a lasso solver built on the primitive operations
used for QR factorizations. When there is only a single solution path, the solver will produce results 
identical to LARS; when there are multiple solution paths, it differs from LARS in that it will produce
a solution path with the smallest active set instead of the largest active set [\citeunique, \S5.1].

Define
$$\eqalign{
\signs(\rg) &\eqdef \fndef{\hsize=3.5in
Set $\A \leftarrow \activer(\rg)$ and return the vector $\s$ such that
$$s_j = \cases{
  0, & \hbox{if $j\le p'$};\cr
  \sign \beta_{\A_j},&\hbox{otherwise}.
}$$
}\cr\noalign{\smallskip}
\complete(\A, \rg) &\eqdef 
\hbox{the vector $\u$ such that
$\u_{-\A}=\nullv$ and $\u_\A = \rg.$
}
}$$
Suppose
$\rghat_0$ minimizes $f(\,\cdot\,;\lambda_0)$ and $\rank \X_\A = \#\A$ where
$\A = \activer(\rghat_0)$. Put $\s\leftarrow\signs(\rghat_0)$,
$$
  \gammab(\lambda) \leftarrow
    \left(\X_\A' \X_\A\right)^{-1}\big[
    \X_\A' \y - \lambda \s
    \big],\quad\hbox{and}\quad
    \xib(\lambda) \leftarrow \X_{-\A}' \big(\y - \X_\A \gammab(\lambda)\big).
$$
If $\min_j \left| \gamma(\lambda)_j\right| \ge 0$ and $\max_j \left| \xi(\lambda)_j\right|\le\lambda$
for all $\lambda \in [\lambda', \lambda_0]$, then Theorem P tells us that
$$
\complete(\A, \gammab(\lambda)) \in \argmin_\rg f(\rg; \lambda)\qquad\hbox{for all $\lambda\in[\lambda', \lambda_0]$}.
$$
The function \step{} takes a piece of a solution path, enlarges its domain as much as possible, and
identifies the constraint that is eventually violated.
$$
\step(\gammab, \xib, \s, \lambda_0) \eqdef \fndef{\hsize=3.5in
Decrease $\lambda'$ starting from $\lambda_0$ until one of the
following conditions is true:
\smallskip
\settabs 20 \columns
\tabalign & $\lambda' = 0$,\cr
\tabalign && return $0$, {\tt NULL};\cr
\smallskip
\tabalign & $\gammab(\lambda')_j = 0$ 
and $\sign \gammab(0)_j \ne \sign s_j$ for some $j>p'$,\cr
\smallskip
\tabalign && return $\lambda'$, $\left({\tt DEACTIVATE},\; j\right)$;\cr
\smallskip
\tabalign& $\left|\xib(\lambda')_j\right| = \lambda'$ and 
${d\over d\lambda} \left|\xib(\lambda)_j\right|$ at $\lambda'<1$,\cr
\smallskip
\tabalign&& return $\lambda'$, $\left({\tt ACTIVATE},\; j\right)$.\cr
}
$$
With \step{}, we can sketch out an algorithm to solve for $\rghat(\lambda)$. Start from $\lambda_0=\infty$ 
and
repeatedly step backwards, solving for segments $\gammab_1, \ldots, \gammab_N$, until the entire interval
$[0, \infty)$ is covered. Before proceeding, though, we should 
address an issue of efficiency: computing $\gammab$ and $\xib$ naively at each step by
evaluating and solving expressions with $\X_\A'\X_\A$ would be expensive.
For a solver to be practical, we need to do better.

Let $\Q_\A$, $\R_\A$ denote a QR factorization for $\X_\A$,
$$
  \X_\A = \Q_\A \left(
  \matrix{
    \R_\A\cr
    \nullv\cr
  }
  \right).
$$
Then
$$
\eqalign{
  \gammab(\lambda) &=
        \left(\X_\A' \X_\A\right)^{-1}
        \left[
          \X_\A' \y - \lambda \s
        \right]\cr
      &= \left(\R_\A' \R_\A\right)^{-1} \left[
        \left(
        \matrix{\R_\A' & \nullv}
        \right)
         \Q_\A' \y - \lambda \s
      \right]\cr
      &= 
      \left(
        \matrix{\R_\A^{-1} & \nullv}
      \right) \Q_\A' \y - \lambda \R_\A^{-1} \R_\A^{-1}~' \s
}$$
and
$$\eqalign{
 \xib(\lambda) &= \X_{-\A}' \left[\y - \X_\A \gammab(\lambda)\right] \cr
 &= 
 \X_{-\A}' \left[
   \y - \Q_\A \left(
   \matrix{
     \R_\A \cr
     \nullv
   }\right)
   \left(
     \left(\matrix{
       \R_\A^{-1} & \nullv
     }\right) \Q_\A' \y - \lambda \R_\A^{-1} \R_\A^{-1}~' \s
   \right)
 \right] \cr
  &= \left(\Q_\A' \X_{-\A}\right)'
  \left[
    \pmatrix{
      \nullv & \nullv \cr
      \nullv & {\bf I}
    }
    \Q_\A' \y
    + \lambda \pmatrix{
      \R_\A^{-1}~' \cr
      \nullv
    }
    \s
  \right].
}$$
Thus, with $\R_\A$, $\Q_\A' \X_{-\A}$, and $\Q_\A'\y$, the
parametric linear equations for $\gammab(\lambda)$ and $\xib(\lambda)$ can be 
computed big-O of
$$
  \big(\hbox{num active regressors}\big)^2 + \big(\hbox{num data}\big) \times \big(\hbox{num inactive regressors}\big).
$$
Let $\F_\A\in\Re^{n\times(p+1)}$ denote what we'll call a {\it factor matrix} for $\A$,
$$
  \F_\A = \pmatrix{
    \matrix{
    \R_\A \cr
    \nullv
    } & \Q_\A' \X_{-\A} & \Q_\A' \y
  }.
$$
If we can update $\F_\A$ as $\A$ changes, we can use it to
efficiently compute $\gammab$ and $\xib$.
We need to consider two cases: 
addition of a new regressor into $\A$ and removal of a regressor from $\A$.
Fortunately, both can be handled by algorithms that mirror
QR factorization by Householder reflectors and Givens rotations, respectively
[\citematrix, ch.~5].
\smallskip
\algbegin Algorithm A (Activate a regressor). Given a factor matrix, $\F$, with $r$ active 
regressors, update $\F$ to activate the $j$-th inactive regressor.

\algstep A1. [Swap columns.] Swap the $(r+1)$-th column of $\F$ with the $(r+j)$-th column.

\algstep A2. [Update $(r+1)$-th Column.] Let $\H$ denote a Householder reflector such
that
$$\displaylines{
\hskip\leftskip\qquad\H
\pmatrix{F_{1,r+1} & \cdots & F_{r, r+1} & * &\cdots & *}'\hfill\cr
\hfill
= \pmatrix{F_{1,r+1} & \cdots & F_{r, r+1} & * & 0 &\cdots & 0}'.\quad\hskip\rightskip\cr
}$$
Set $\F \leftarrow \H \F$.\qed
\medskip
\noindent Algorithm A can be implemented using the LAPACK routines {\tt LARFG} and {\tt ORMQR}, and it
has computational complexity big-O of
$$
\left(\hbox{num data}\right) + \left(\hbox{num data} - \hbox{num active regressors}\right) \times \left(\hbox{num inactive regressors}\right).
$$
\smallskip
\algbegin Algorithm D (Deactivate a regressor). Given a factor matrix, $\F$, with $r$ active regressors,
update $\F$ to deactivate the $j$-th active regressor.

\algstep D1. [Permute Columns.] Pick ${\bf P}$ to be a permutation matrix so that ${\bf P} \F$ performs
a left cyclical rotation of columns $j$, $j+1$, $\ldots$, $r$ of $\F$:
$$
\left({\bf P}\F\right)_j = \F_{j+1},\quad\ldots,\quad
\left({\bf P}\F\right)_{r-1} = \F_r, \quad\hbox{and}\quad
\left({\bf P}\F\right)_r = \F_j. 
$$
Set $\F \leftarrow {\bf P} \F$.

\algstep D2. [Annihilate subdiagonal.] For $j \le k < r$, let $\G_k$ denote a Givens rotation such that
$$\displaylines{
\hskip\leftskip\qquad\G_k 
\pmatrix{F_{1k} & \cdots & F_{k-1, k} & * & * & 0 & \cdots & 0}'\hfill\cr
\hfill
= \pmatrix{F_{1k} & \cdots & F_{k-1, k} & * & 0 & 0 & \cdots & 0}'.\quad\hskip\rightskip\cr
}$$
Set $\F \leftarrow \G_{r-1} \G_{r-2}\cdots\G_j \F$.\enspace\qed
\medskip
\noindent Similarly, Algorithm D can be implemented with the BLAS routine
{\tt ROT} and has computational cost big-O of
$$
\left(\hbox{num active regressors}\right)^2 + \left(\hbox{num active regressors}\right) \times
\left(\hbox{num inactive regressors}\right).
$$

With Algorithm A and Algorithm D, we are now in a position to put together an efficient solver 
for $\rghat(\lambda)$.
\smallskip
\algbegin Algorithm S (Compute a full rank solution path). Suppose
$\X\in\Re^{n\times p}$, $\y \in \Re^n$, and
$
\rank \X_{\left\{ 1, \ldots, p'\right\}} = p'.
$
Produce a piecewise linear function
$\rghat(\lambda)$ such that for any $\lambda \ge 0$,
$$
\rghat(\lambda) \in \argmin_{\rg}
\left\{
  {1\over2}
     \left\|\y - \X \rg\right\|^2 + \lambda \sum_{j=p'+1}^p \left| \beta_j\right|
\right\}
$$
and the nonzero entries of $\rghat(\lambda)$ identify regressors of full rank.
\algstep S1. [Initialize.] Set
$\A \leftarrow \left(1, \ldots, p'\right)$,\enspace\enspace$\s \leftarrow \nullv_{p'}$, \enspace\enspace
$\lambda_1\leftarrow\infty$, \enspace and $k\leftarrow 1$. Initialize $\F$ to a factor 
matrix for $\A$ using Algorithm A and put
$$
\rg_1 \leftarrow \complete\left(\A, \;\R_\A^{-1}\Q_\A'\y\right).
$$

\algstep S2. [Loop $k$.] Until $\lambda_k=0$, do Step S3 and set $k\leftarrow k+1$. When
done, go to Step S4.

\algstepbegin S3. [Step solution.] Extract $\R_\A$, $\Q_\A' \X_{-\A}$, and $\Q_\A' \y$ from 
the factor matrix $\F$. Set
$$\eqalign{
  \gammab(\lambda) &\leftarrow
      \left(
        \matrix{\R_\A^{-1} & \nullv}
      \right) \Q_\A' \y - \lambda \R_\A^{-1} \R_\A^{-1}~' \s,\cr
 \xib(\lambda) 
  &\leftarrow \left(\Q_\A' \X_{-\A}\right)'
  \left[
    \pmatrix{
      \nullv & \nullv \cr
      \nullv & {\bf I}
    }
    \Q_\A' \y
    + \lambda \pmatrix{
      \R_\A^{-1}~' \cr
      \nullv
    }
    \s
  \right],\cr
  \lambda_{k+1},\;{\tt ACT} &\leftarrow \step(\gammab, \xib, \s, \lambda_k),
  \quad\hbox{and}\quad\rg_{k+1}\leftarrow \complete(\A, \gammab(\lambda_{k+1})).
}$$
If ${\tt ACT}=({\tt ACTIVATE},\,j)$, add the
regressor for $j$ to $\A$, append $\sign \xib(\lambda_{k+1})_j$ to $\s$, and update $\F$ 
using Algorithm A.
\smallskip
If
${\tt ACT}=({\tt DEACTIVATE}, \, j)$, remove the $j$-th entry from $\A$ and
$\s$. Then update $\F$ using Algorithm D.
\smallskip
Return to Step S2.
\algstepend

\algstep S4. [Assemble solution.] Return $\rghat(\lambda)$, the piecewise linear function
that goes through the nodes
$
(\lambda_k, \rg_k), \ldots, (\lambda_2, \rg_2)
$
and equals $\rg_1$ for $\lambda > \lambda_2$.

\subend

\medskip
\ifx\mIncludeGuard\undfined
\input mac/m-impl
\def\subend{\end}
\else
\def\subend{}
\fi

\newsec{LO-LARS Algorithm} In this section, we adapt Algorithm S
to compute leave-one-out error paths and the LO function.

Given a solution path, $\rghat(\lambda)$, define
$
T_\rghat(\lambda) \eqdef \sum_{j=p'+1}^p | \rghat(\lambda)_j |
$.
$T_\rghat$ computes the partial $\ell_1$ norm of a solution. Since $\rghat$ is a
piecewise parametric linear function of $\lambda$ where the sign of a restricted regressor
is constant on
any segment, $T_\rghat$ is also a piecewise linear function of $\lambda$ with the same
segments as $\rghat$.

Suppose $B=(\rghat\idx{-1}(\lambda), \ldots, \rghat\idx{-n}(\lambda))$ 
is an array of solution
paths such that $\rghat\idx{-i}(\lambda)$ minimizes 
$$
  f(\rg; \lambda)
   - \half\left(y\idx{i} - \x\idxp{i} \rg\right)^2.
$$
Here are two reasonable ways we might define LO for $B$:
$$
\hbox{LO}\idx1(\lambda; B) = 
    \sum_{i=1}^n \left(y_i - \rghat\idx{-i}(\lambda)\right)^2
    \quad\hbox{or}\quad
\hbox{LO}\idx2(t; B) = 
    \sum_{i=1}^n \left(y_i -
    \rghat\idx{-i}(T^{-1}_{\rghat\idx{-i}}(t)) \right)^2.
$$
$\hbox{LO}\idx1$ keeps the penalty multiplier $\lambda$ constant across leave-one-out
subproblems, and $\hbox{LO}\idx2$ keeps the $\ell_1$ restriction constant across
leave-one-out subproblems.

We can sketch out a high level algorithm to compute 
$\hbox{LO}\idx1$ and $\hbox{LO}\idx2$.
\smallskip
\algbegin Algorithm L (Compute LO). Given
$\X\in\Re^{n\times p}$ and $\y \in \Re^n$, compute $\hbox{LO}\idx1$ and 
$\hbox{LO}\idx2$ for some collection
of leave-one-out solution paths $B$.

\algstep L1. [Produce error paths.] For each $i$, compute the segments of a piecewise linear
leave-one-out error function,
$$
  \left(\lambda^{(i)}_1, t^{(i)}_1, e^{(i)}_1\right), \ldots,
  \left(\lambda^{(i)}_{m_i}, t^{(i)}_{m_i}, e^{(i)}_{m_i}\right),\qquad 1\le i \le n
$$
where $\infty=\lambda^{(i)}_1 > \cdots > \lambda^{(i)}_{m_i} = 0$ and for some leave-one-out
solution path $\rghat\idx{-i}$,
$$\displaylines{
\hbox{$\rghat\idx{-i}(\lambda)$ is linear on $[\lambda^{(i)}_k, \lambda^{(i)}_{k+1}]$},\quad
t^{(i)}_k = \sum_{j=p'+1}^p |\rghat\idx{-i}(\lambda^{(i)}_k)_j|,\quad\hbox{and}\cr
e^{(i)}_k = y\idx{i} - \x\idxp{i}\,\rghat\idx{-i}(\lambda^{(i)}_k).
}$$

\algstep L2. [Assemble LO.] Let $e\idx{i}(\lambda)$ denote the piecewise linear function that
connects the points $\{(\lambda^{(i)}_k, e^{(i)}_k)\}$ and let $B$ denote the collection of
solution paths from Step L1. Then
$$
  \hbox{LO}^{(1)}(\lambda; B) = \sum_{i=1}^n e\idx{i}(\lambda)^2.
$$
Similarly, we can connect the piecewise linear points 
$\{(t\idx{i}_k, e\idx{i}_k)\}$ to compute $\hbox{LO}\idx2$.\qed
\smallskip
\noindent To make Step L1 tractable, we need some of the computation of individual error paths
to be shared.

Suppose 
$$\eqalign{
\gammab\idx{-i}(\lambda) &= 
\left( \X\idxp{-i}_{\A} \X\idx{-i}_{\A}\right)^{-1}\left[
\X\idxp{-i}_{\A} \y\idx{-i} - \lambda \s
\right]\cr
&= \left(\X_{\A}' \X_{\A} - \x\idx{i}_\A \x\idxp{i}_\A\right)^{-1}
\left[
\X_{\A}' \y - y\idx{i} \x\idx{i}_\A - \lambda \s
\right]
}$$
corresponds to a segment of a solution for the $i$-th leave-one-out lasso subproblem.
Applying the matrix inversion lemma,
$$
\left(\Amat + \u \v'\right)^{-1} = 
\Amat^{-1} - {\Amat^{-1}\u\v'\Amat^{-1}\over 1 + \v' \Amat^{-1}\u},
$$
gives us
$$\eqalign{
\left(\X_{\A}' \X_{\A} - \x\idx{i}_\A \x\idxp{i}_\A\right)^{-1}
  &= 
  \left(\X_{\A}' \X_{\A}\right)^{-1}
  + {
  (\X_{\A}'\X_{\A})^{-1} \x\idx{i}_\A
  \x\idxp{i}_\A(\X_{\A}'\X_{\A})^{-1}
  \over
  1 - 
  \x\idxp{i}_\A (\X_{\A}'\X_{\A})^{-1} \x\idx{i}_\A}\cr
  &= 
  \left(\X_{\A}' \X_{\A}\right)^{-1}
  + {\w\idx{i}_\A \w\idxp{i}_\A\over{1 - h\idx{i}_\A}}
}\filleqno{lo}$$
with
$\w\idx{i}_\A = 
  (\X_{\A}' \X_{\A})^{-1}\x\idx{i}_\A$ and
$h\idx{i}_\A =
  \x\idxp{i}_\A (\X_{\A}' \X_{\A})^{-1}\x\idx{i}_A$.
Thus,
$$
  \gammab\idx{-i}(\lambda) = 
    \gammab\idx{-i}(0) + \lambda\,\dot{\gammab}\idx{-i}(0)
$$
where
$$\eqalign{
    \gammab\idx{-i}(0)
      &= \left[
      \left(\X_\A' \X_\A\right)^{-1} + 
      {\w\idx{i}_\A \w\idxp{i}_\A\over1 - h\idx{i}_\A}
      \right]\left(
      \X_\A' \y - y\idx{i}\x\idx{i}_\A
      \right)\cr
      &= 
      \gammab(0)
      + 
      {\w\idx{i}_\A \w\idxp{i}_\A\over1 - h\idx{i}_\A} \X_\A' \y
      - \left(\X_\A' \X\right)^{-1} y\idx{i}\x\idx{i}_\A
      - {\w\idx{i}_\A \w\idxp{i}_\A\over1 - h\idx{i}_\A} y\idx{i} \x\idx{i}_\A \cr
      &= 
      \gammab(0) +
      {\x\idxp{i}_\A \gammab(0)
      \over1 - h\idx{i}_\A} \w\idx{i}_\A
      - y\idx{i} \w\idx{i}_\A
      - {h\idx{i}_\A y\idx{i} \over1 - h\idx{i}_\A} \w\idx{i}_\A\cr
      &= 
      \gammab(0)
    -
      {y\idx{i} - \x\idxp{i}_\A \gammab(0)
      \over1 - h\idx{i}_\A} \w\idx{i}_\A
}$$
and
$$
\dot{\gammab}\idx{-i}(0)
      = -\left[
      \left(\X_\A' \X_\A\right)^{-1} + 
      {\w\idx{i}_\A \w\idxp{i}_\A\over1 - h\idx{i}_\A}
      \right] \s
      = 
      \dot{\gammab}(0)
      - {\w\idxp{i}_\A \s\over1 - h\idx{i}_\A} \w\idx{i}_\A.
$$
We can then express $\gammab\idx{-i}$ in terms of an adjustment to $\gammab$,
$$
\gammab\idx{-i}(\lambda) = 
  \gammab(\lambda) + 
  \delta\gammab\idx{i}(\lambda)
$$
where
$$\displaylines{
\delta\gammab\idx{i}(\lambda) =
-{y\idx{i} - \x\idxp{i}_\A\gammab(0)\over1 - h\idx{i}_\A} \w\idx{i}_\A
- \lambda\,{\w\idxp{i}_\A \s\over1-h\idx{i}_\A} \w\idx{i}_\A
= a_\A\idx{i}(\lambda) \w\idx{i}_\A\cr
\hbox{with}\quad a\idx{i}_\A(\lambda)
= 
-{y\idx{i} - \x\idxp{i}_\A\gammab(0)\over1 - h\idx{i}_\A} 
- \lambda\,{\w\idxp{i}_\A \s\over1-h\idx{i}_\A}.\cr
}$$
Similarly, we have
$$\eqalign{
 \xib\idx{-i}(\lambda) 
 &= \X_{-\A}\idxp{-i} \left[ \y\idx{-i} - \X_\A\idx{-i} \gammab\idx{-i}(\lambda) \right]\cr
 &= \X_{-\A}' \left[\y - \X_\A \gammab\idx{-i}(\lambda)\right]
 -\left[y\idx{i} - \x\idxp{i}_\A \gammab\idx{-i}(\lambda)\right]
 \x\idx{i}_{-\A} 
 \cr
 &= \xib(\lambda) + \delta\xib\idx{i}(\lambda)\cr
}$$
with
$$
 \delta\xib\idx{i}(\lambda) 
 = - \X'_{-\A} \X_\A\delta\gammab\idx{i}(\lambda)
 -\left[y\idx{i} - \x\idxp{i}_\A \gammab\idx{-i}(\lambda)\right]
 \x\idx{i}_{-\A}.
$$
If $\Q_\A$ and $\R_\A$ denote the QR factorization of $\X_\A$, then we can write the
equations as
$$\displaylines{
  h\idx{i}_\A = \left\| \R_\A^{-1\prime} \x\idx{i}_\A \right\|^2,\qquad
  \w\idx{i}_\A = \R_\A^{-1}\R_\A^{-1\prime} \x\idx{i}_\A,\quad\hbox{and}\cr
  \delta\xib\idx{i}(\lambda) =
-a\idx{i}_\A(\lambda) \left(\Q_\A'\X_{-\A}\right)'
\pmatrix{
\R_\A^{-1\prime}\cr
\nullv\cr
}\x\idx{i}_\A 
 -\left[y\idx{i} - \x\idxp{i}_\A \gammab\idx{-i}(\lambda)\right]
 \x\idx{i}_{-\A}.\cr
}$$

The above equations show that $\gammab\idx{-i}$ and $\xib\idx{-i}$ be computed as updates
to $\gammab$ and $\xib$. And given the factor matrix $\F_\A$, $\gammab$, and $\xib$,
the leave-one-out functions $\gammab\idx{-i}$ and $\xib\idx{-i}$ can be computed big-O of
$$
\big(\hbox{num active regressors}\big)
\times \big(\hbox{num regressors}\big).
$$
Thus, when leave-one-out path segments share the same set of active regressors,
they can be computed with much lower cost than if treated independently. To capitalize
on this observation, we can maintain a computation cache that maps a set of active
regressors to a shared factor matrix. A natural data structure to use is a hash map indexed
with bitsets to represent active regressors.
Then we can maintain a second
level of caching that maps a pair $(\A, \s)$ to common 
base parametric equations $\gammab$ and $\xib$ where $\A$ is a vector of active regressors and
$\s$ is a sign vector. We can access and update such a cache with the functions
$$
  {\tt lookup}({\tt CACHE}, \A, \s) \eqdef \fndef{\hsize=3.3in
  Let $(\A', \s')$ denote a paired active regressor and sign vector in {\tt CACHE}
  that matches $(\A, \s)$ up to a reordering.
  \smallskip
  Let $\F_{\A'}$ denote the cached factor matrix for $\A'$ and let 
  $(\gammab_{\A', \s'},\; \xib_{\A', \s'})$ denote
  the cached parametric equations for $(\A', \s')$ as defined in Step 3 of Algorithm S.
  \smallskip
  Return $\left(\A', \;\s', \;\F_{\A'}, \;\gammab_{\A', \s'}, \;\xib_{\A', \s'}\right)$.
  }
$$
and
$$\displaylines{
\hskip\leftskip\quad{\tt update}({\tt CACHE}, \A, \s, \F_{\A}, {\tt ACT}, s') \eqdef\hfill\cr
\noalign{\vskip-10pt}
\hfill
\quad\cr
  \fndef{\hsize=4.3in
  If ${\tt ACT} = ({\tt ACTIVATE}, j)$, then add the
  regressor for $j$ to $\A$ and append $s'$ to $\s$.
  If there is not already an entry for $\A$ in {\tt CACHE}, copy $\F_\A$ if necessary,
  update using Algorithm A, and store in {\tt CACHE}.
  \smallskip
  If ${\tt ACT} = ({\tt DEACTIVATE}, j)$, remove the $j$-th entry from $\A$ and $\s$.
  If there is not already an entry for $\A$ in {\tt CACHE}, copy $\F_\A$ if necessary,
  update using Algorithm D, and store in {\tt CACHE}.
  \smallskip
  Rearrange $\A$ and $\s$ to match the ordering used for its cached factor matrix. If
  there is not already an entry for the pair $(\A, \s)$, compute 
  $(\gammab_{\A, \s},\;\xib_{\A, \s})$ using the equations from Step 3 of Algorithm S
  and store in {\tt CACHE}.
  \smallskip
  Return $(\A, \s)$.
  }\cr
}$$
Let's now flesh out Step L1 in greater detail.
\smallskip
\algbegin Algorithm E (Compute leave-one-out error segments). Given
$\X\in\Re^{n\times p}$ and $\y \in \Re^n$ such that
$
\rank \X_{\left\{1, \ldots, p'\right\}} = p'
$,
compute the leave-one-out error segments,
$$
  \left(\lambda^{(i)}_1, t^{(i)}_1, e^{(i)}_1\right), \ldots,
  \left(\lambda^{(i)}_{m_i}, t^{(i)}_{m_i}, e^{(i)}_{m_i}\right),\qquad 1\le i \le n
$$
described in Algorithm L.

\algstep E1. [Initialize.] Set
$\A_0 \leftarrow \left(\,1,\ldots,p'\right)'$ and $k \leftarrow 1$. Compute a factor
matrix $\F_{\A_0}$ for $A_0$ using Algorithm A. Initialize ${\tt CACHE}$ to be empty; then
add an entry for the pair $(\A_0, \F_{\A_0})$. 
For $i=1, \ldots, n$, set
$$
\A\idx{i}\leftarrow\A_0,\quad \s\idx{i}\leftarrow\nullv_{p'},
\quad
\lambda\idx{i}_1 = \infty,
\quad\hbox{and}\quad 
t\idx{i}_1 \leftarrow 0.
$$

\algstep E2. [Main loop.] Until $\lambda\idx{i}_k = 0$ for all $i$, do Step E3 and set 
$k\leftarrow k + 1$.
\algstep E3. [Loop $i$.] Initialize ${\tt CACHE}'$ to be empty. 
For $i=1,\ldots,n$, do Step E4. When done, set 
${\tt CACHE} \leftarrow {\tt CACHE}'$ and return to Step E2.
\algstepbegin E4. [Extend error path.] 
If $\lambda\idx{i}_k = 0$, return to Step E3. Otherwise, set
$$
\A,\; \s,\;\F_{\A},\; \gammab, \;\xib
\leftarrow {\tt lookup}({\tt CACHE}, \A\idx{i}, \s\idx{i}).
$$
Extract $\R_\A$ and $\Q_\A'\X_{-\A}$ from $\F_\A$.
\smallskip

Set $\u \leftarrow \R_\A^{-1\prime} \x\idx{i}_\A$, $h \leftarrow \|\u\|^2$,
$\w \leftarrow \R_\A^{-1}\u$. Then put
$$\displaylines{
a(\lambda) \leftarrow
-{y\idx{i} - \x\idxp{i}_\A\gammab(0)\over1 - h}
- \lambda\,{\w' \s\over1-h},\qquad
\gammab\idx{-i}(\lambda) \leftarrow  \gammab(\lambda) + a(\lambda)\w,\quad\hbox{and}\cr
\xib\idx{-i}(\lambda) \leftarrow
\xib(\lambda)
  - a(\lambda) \left(\Q_{\A}' \X_{-\A}\right)' 
  \pmatrix{
  \u\cr
  \nullv\cr
  }
 -\left[y\idx{i} - \x\idxp{i}_\A \gammab\idx{-i}(\lambda)\right]
 \x\idx{i}_{-\A}.
}$$
\smallskip
Put $\lambda\idx{i}_{k+1}, {\tt ACT}\leftarrow \step(\gammab\idx{-i},
\xib\idx{-i}, \s, \lambda\idx{i}_k)$
and $\rghat \leftarrow \gammab\idx{-i}(\lambda\idx{i}_{k+1})$.
\smallskip
If $k=1$, set $e\idx{i}_1 \leftarrow 
y\idx{i} - \x\idxp{i}_\A \left(\gammab\idx{-i}(0)\right)$.
\smallskip
Set $e\idx{i}_{k+1}\leftarrow y\idx{i} - \x\idxp{i}_\A \rghat$ and
$t\idx{i}_{k+1}\leftarrow \sum_{j=p'+1}^p |\hat{\beta}_j|$.
\smallskip
If ${\tt ACT} = ({\tt ACTIVATE}, j)$, set $s' = \sign \xib\idx{-i}(\lambda\idx{i}_{k+1})_j$;
otherwise, set $s'=0$.
\smallskip
If ${\tt ACT} \ne {\tt NULL}$, invoke 
$$
\A\idx{i},\;\s\idx{i} \leftarrow 
  {\tt update}({\tt CACHE}', \A, \s, \F_{\A}, {\tt ACT}, s').
$$
Return to Step E3.\qed

\algstepend
\smallskip
\noindent The cost of Algorithm E depends heavily on the extent of divergence in the leave-one-out solution
paths. The more individual solution paths overlap in their active regressors, the more
efficient Algorithm E will be. We can make several modifications to improve performance.
\smallskip
\noindent{\it Update batching.} As multiple solution paths can share the same factor matrix, we need to ensure 
that
we don't modify a factor matrix that's used by another path. We can copy factor matrices before updating,
 but that leads to additional overhead. However, if we rework Step E4 so that the updates are done as a batch
 at the end of the loop in Step E3, then we can elide the unnecessary copies.
\smallskip
\noindent{\it Early stopping.} With high-dimensional data sets, solution paths can diverge 
substantially when there is little shrinkage (i.e.\ $t$ is large).
In such cases, it can be beneficial to modify Algorithm L so that instead of
computing the entire LO function, Algorithm L exits early if LO exceeds a specified
percentage of the best optimum found so far. We can make this possible by merging Step L1
with Step L2 so that the LO curve is built out as segments become available. We then track the
best local optimum of the curve and regularly monitor LO to see if the exit condition is met.
\subend

\medskip
\ifx\mIncludeGuard\undfined
\input mac/m-impl
\def\subend{\end}
\else
\def\subend{}
\fi

\newsec{LO-LARS for ALO} With some adjustments, we can rework Algorithm E to compute ALO.
Let $\rghat(\lambda)$ denote a solution path. Given $\lambda$,
ALO approximates LO by
$$\displaylines{
\hbox{LO for $\lambda$} \approx \sum_i e\idx{i}(\lambda)^2\quad\hbox{where}\cr
e\idx{i}(\lambda) = 
y\idx{i} - \x\idxp{i}_\A \left(\X\idxp{-i}_\A \X\idx{-i}_\A\right)^{-1}
\left[
\X\idxp{-i}_\A \y\idx{-i} - \lambda \s
\right],\cr
\s = \signs(\rghat(\lambda)),\quad\hbox{and}\quad \A = \activer(\rghat(\lambda)). \cr
}$$
Applying the derivations from (\eqlo) gives us
$$\eqalign{
e\idx{i}(\lambda) &=
y\idx{i} - \x\idxp{i}_\A 
\left[
\left(\X_\A' \X_\A\right)^{-1}
+ {\w\idx{i}_\A \w\idxp{i}_\A\over1 - h\idx{i}_\A}
\right]
\left[
\X_\A' \y - y\idx{i} \x\idx{i}_\A - \lambda \s
\right]\cr
&= \left(y\idx{i} - \x\idxp{i} (\X_\A'\X_\A)^{-1}\X_\A' \y\right)  
{1\over1 - h\idx{i}_\A} +
\lambda {1\over1 - h\idx{i}_\A} \w\idxp{i}_\A \s.
}$$
Using these equations, we can write a simplified version of Algorithm E for ALO. Note, though,
that unlike LO, ALO can have jump discontinuities; so when computing ALO about a point
$\lambda$ where the activation changes, we need to provide values both for the left-hand limit
and the right-hand limit.
\smallskip
\algbegin Algorithm X (Compute approximate leave-one-out error segments). Given
$\X\in\Re^{n\times p}$ and $\y \in \Re^n$ such that
$
\rank \X_{\left\{1, \ldots, p'\right\}} = p'
$,
compute the approximate leave-one-out error segments,
$$
  \left(\lambda_1, \l_1, \r_1\right), \ldots,
  \left(\lambda_{m}, \l_m, \r_m\right),
$$
such that
$$
\hbox{ALO for $\lambda_k < \lambda < \lambda_{k+1}$} =
\sum_{i=1}^n \left[r_k\idx{i} {\lambda - \lambda_{k+1}\over\lambda_k - \lambda_{k+1}}
+ l_{k+1}\idx{i} {\lambda - \lambda_k\over\lambda_{k+1}-\lambda_k}
\right]^2.
$$

\algstep X1. [Initialize.] Set
$\A \leftarrow \left(1, \ldots, p'\right)$,\enspace\enspace$\s \leftarrow \nullv_{p'}$, \enspace\enspace
$\lambda_1\leftarrow\infty$, \enspace and $k\leftarrow 1$. Initialize $\F$ to a factor 
matrix for $\A$ using Algorithm A.

\algstep X2. [Loop $k$.] Until $\lambda_k=0$, do Step X3 to Step X5 and set $k\leftarrow k+1$.

\algstepbegin X3. [Step solution.] Extract $\R_\A$, $\Q_\A' \X_{-\A}$, and $\Q_\A' \y$ from 
the factor matrix $\F$. Set
$$\displaylines{
  \gammab(\lambda) \leftarrow
      \left(
        \matrix{\R_\A^{-1} & \nullv}
      \right) \Q_\A' \y - \lambda \R_\A^{-1} \R_\A^{-1}~' \s,\cr
 \xib(\lambda) 
  \leftarrow \left(\Q_\A' \X_{-\A}\right)'
  \left[
    \pmatrix{
      \nullv & \nullv \cr
      \nullv & {\bf I}
    }
    \Q_\A' \y
    + \lambda \pmatrix{
      \R_\A^{-1}~' \cr
      \nullv
    }
    \s
  \right],\cr
  \lambda_{k+1},\;{\tt ACT} \leftarrow \step(\gammab, \xib, \s, \lambda_k),\cr
  \v\leftarrow \R_\A^{-1}~' \s, 
  \quad\hbox{and}\quad \varepsilon\leftarrow 
  y\idx{i} - \X_\A\idxp{i} \gammab(0).
}$$

\algstepend
\algstepbegin X4. [Loop $i$.] For $i=1,\ldots,n$, set
$$\displaylines{
\u\idx{i} \leftarrow \R_\A^{-1}~' \x_\A\idx{i},\qquad
h\idx{i} \leftarrow \| \u\idx{i}\|^2,\qquad \alpha\idx{i} \leftarrow {\varepsilon\over 1 - h\idx{i}},
\qquad \beta\idx{i} \leftarrow 
{1\over1-h\idx{i}} \v' \u\idx{i},
}$$
and put
$
r_k\idx{i} \leftarrow \alpha\idx{i} + \lambda_{k} \beta\idx{i}$
and
$
l_{k+1}\idx{i} \leftarrow \alpha\idx{i} + \lambda_{k+1} \beta\idx{i}$.
If $k = 1$, set $l_k\idx{i} \leftarrow r_k\idx{i}$.

\algstepend
\algstepbegin X5. [Update factor matrix.]
If ${\tt ACT}=({\tt ACTIVATE},\,j)$, add the
regressor for $j$ to $\A$, append $\sign \xib(\lambda_{k+1})_j$ to $\s$, and update $\F$ 
using Algorithm A.
\smallskip
If
${\tt ACT}=({\tt DEACTIVATE}, \, j)$, remove the $j$-th entry from $\A$ and
$\s$. Then update $\F$ using Algorithm D.
\smallskip
Return to Step X2.\qed

\algstepend
\medskip
\noindent The most expensive part of Algorithm X is this computation in Step X4:
$$
\u\idx{i} \leftarrow \R_\A^{-1}~' \x_\A\idx{i}.
$$
When done for all $i$, the cost is big-O of
$
\left(\hbox{num data}\right)\times\left(\hbox{num active regressors}\right)^2
$. But with some additional bookkeeping, we can reduce the cost substantially.
Suppose that Step X5 activates a regressor. Let $\A$ denote the active regressors before
Step X5 and let $\A'$ denote the active regressors after. Then the matrix $\R_{\A'}$ has the
form
$$
\R_{\A'} = 
\pmatrix{
\R_\A & \r_1 \cr
\nullv & r_2 \cr
}
$$
so that
$$
\R_{\A'}^{-1}~' \x_{\A'}\idx{i}
  = 
\pmatrix{
 \R_\A' & \nullv\cr
 \r_1' & r_2\cr
}^{-1} \x_{\A'}\idx{i}
  = 
\pmatrix{
 \R_\A^{-1}~' & \nullv\cr
 -r_2^{-1} \r_1'\R_\A^{-1}~' & r_2^{-1} \cr
} \x_{\A'}\idx{i}.
$$
Thus, we see that $\R_{\A'}^{-1}~'\x_{\A'}\idx{i}$ can be computed as an update to
$\R_{\A}^{-1}~'\x_\A\idx{i}$ with cost ${\cal O}(\hbox{num active regressors})$.
Similarly, a deactivation at index $j$ can be computed by deactivating the last 
$\#\A - j + 1$ regressors and then reactivating regressors $j+1$ to $\#\A$ so that
the cost is big-O of
$$
\left(\hbox{num active regressors}\right)\times\left(\hbox{num active regressors} - \hbox{deactivation index}\right).
$$

\medskip
\ifx\mIncludeGuard\undfined
\input mac/m-impl
\def\subend{\end}
\else
\def\subend{}
\fi

\newsec{Numerical Experiments} Consider a linear model
$$
y = \x' \rgtrue + \varepsilon,\qquad \x\sim N(\nullv_p, \Sigma),\quad\hbox{and}\quad
\varepsilon \sim N(0, \sigma^2).
$$
Suppose $\rghat$ denotes an estimator for the parameter $\rgtrue$ where $\rghat$ maps
sample observations $(\x_1, y_1), \ldots, (\x_n, y_n)$ to an estimated regressor.
For a given $\rg$, we will have an expected out-of-sample 
prediction error of
$$\eqalign{
\pe(\rg) &= {\bbb E}\left[\left(\x' (\rg - \rgtrue ) - \varepsilon\right)^2\mid\rg\right]\cr
  &=
  {\bbb E}\left[
  (\rg - \rgtrue)'\x\x'(\rg - \rgtrue) -
  2 \varepsilon \x' (\rg - \rgtrue) + \varepsilon^2\mid\rg
  \right]\cr
  &= (\rg - \rgtrue)'\Sigma(\rg-\rgtrue) + \sigma^2.
}$$
Now suppose that we estimate the prediction error of $\rghat$ with ${\widehat {\rm pe}}$ using the
same observations. Then the bias and variance of ${\widehat {\rm pe}}$
are given by
$$
  {\bbb E}\left[\widehat {\rm pe} - \rm{pe}(\rghat)\right]\quad\hbox{and}\quad
  {\bbb V}\left[\widehat {\rm pe} - \rm{pe}(\rghat)\right].
$$
Perhaps the simplest ${\widehat {\rm pe}}$ is the holdout error estimator, where one fraction of the
data is used for training and the other fraction is used for error estimation. 
We can expect holdout to have a bias as the holdout training data set is smaller than the data set used to fit 
the actual model. For the case of OLS, we can work out a simple equation for the bias. Letting
$\rghat^{\rm OLS}$ denote the OLS estimator and setting $\e \leftarrow \y - \X \rgtrue$, we have
$$\eqalign{
\rghat^{\rm OLS} - \rgtrue &= \left(\X'\X\right)^{-1}\X' \e,\cr
\pe(\rgols) &= \e' \X \left(\X'\X\right)^{-1} \Sigma \left(\X'\X\right)^{-1} \X' \e + \sigma^2\cr
    &= \tr\left[\left(\X'\X\right)^{-1} \Sigma \left(\X'\X\right)^{-1} \X' \e\e' \X\right] + \sigma^2,\quad\hbox{and}\cr
{\bbb E}\left[\pe(\rgols)\mid \X\right] 
    &= \sigma^2 \tr\left[\left(\X'\X\right)^{-1} \Sigma \right] + \sigma^2.\cr
}$$
Now $(\X'\X)^{-1}$ follows an inverse Wishart distribution with $n$ degrees of freedom [\citewishart, \S7.7] and has expected value
$
{\bbb E}\left[(\X'\X)^{-1}\right] = {\Sigma^{-1} \over n - p - 1}
$.
Thus,
$$
{\bbb E}\left[\pe(\rgols)\right] = {\sigma^2 p \over n - p - 1} + \sigma^2.
$$
So if we estimate the prediction error for a data set of $n$ observations using holdout with a training set of size $n'$,
the bias will be
$$
{\sigma^2 p\over n'-p -1} - {\sigma^2p\over n - p -1}
= \sigma^2 p {n - n' \over (n' - p - 1)(n - p -1)}.
$$
Clearly, bias decreases with $n'$ closer to $n$. But also interesting to note is the role of $p$: with $p$ close to $n$,
a larger training set size becomes more important for reducing bias and even a slightly larger size can lead to a substantial
reduction in bias.

Cross-validation averages over multiple holdout estimates. While cross-validation
won't improve on the bias of the holdout estimators (assuming the folds are of equal size), it can reduce the variance significantly.
We can use Monte Carlo simulations to explore how well LO and other forms of $v$-fold cross-validation perform
as estimators for the lasso.
\smallskip
\noindent{\bf Simulation B.}
For our first simulation, we'll set $n=50$, $p=40$, and  assume the covariance matrix $\Sigma$ has a Toeplitz structure with
$
  \Sigma_{ij} = 0.5^{|i - j|}
$. We set
$$
\left(\rgtrue\right)_j = \cases{
  {1\over20},\quad\hbox{if $j\le 20$;}\cr\noalign{\smallskip}
  \;0,\quad\hbox{otherwise;}\cr
}
\quad\hbox{and}\quad \sigma^2 =
\rgtrue' \Sigma \rgtrue = 0.14
$$
so that the model has a signal-to-noise ratio of $1$. The table below shows the performance
of $v$-fold cross-validation for various values of $v$ and $t$ over $100{,}000$ instances.
\displaytable{.1in}{table/sim1}{}{%
Results of a Monte Carlo simulation to measure the bias and variance of $v$-fold cross-validation as an
estimator for lasso regression prediction error for $v=5, 10, n$ and various values of the 
shrinkage parameter $t$. Also shown for each $t$ are the average number of active regressors and 
the average prediction error.
}
As we would expect, the bias of LO is notably lower than the bias of 5-fold and 10-fold cross-validation. And like OLS,
the differences in the biases are more pronounced with more active regressors.
The differences in the variances are all minor.
\smallskip
For the next simulation, we investigate the accuracy of estimating the optimal value of $t$ in terms of prediction
error by optimizing LO.
\smallskip
\noindent{\bf Simulation O.}
We use the same values for $n$, $p$, and $\Sigma$ as Simulation B. For each simulated data set, an oracle
value $t_{\rm opt}$ is selected so as to optimize the true prediction error, 
$$
t_{\rm opt} = \argmin_t \left\{{\rm pe}(\rghat(t))\right\}, 
$$
and an estimated value $\hat{t}_{\rm opt}$ is selected so as to optimize LO. The table below summarizes 
the performance of the estimator over $10{,}000$ instances.
\displaytable{.1in}{table/sim2}{}{%
Results of a Monte Carlo simulation to measure the accuracy of using the LO minimizer to estimate
the prediction error minimizer. Also shown are the average number of active regressors, the average
prediction error of the LO minimum, and the average actual prediction error minimum.
}
While $\hat{t}_{\rm opt}$ performs decently, 
it does appear to be notably larger than the true\hfil\break
value, suggesting that there might be a better way to estimate the optimum.
\subend

\medskip
\ifx\mIncludeGuard\undfined
\input mac/m-impl
\def\subend{\end}
\else
\def\subend{}
\fi

\newsec{Conclusions} LO can be an effective way to estimate prediction error for the lasso.
It has lower bias than other forms of $v$-fold cross-validation, particularly for 
high-dimensional data sets; and as we saw in Simulation B, the variance of LO can be
nearly identical to that of $10$-fold and $5$-fold cross-validation.
The LO-LARS algorithm makes exact computation of LO and ALO practical for many real-world data sets
and enables fully deterministic and reproducible selection of the lasso shrinkage
parameter.

While I think LO is an excellent default choice for estimating prediction error,
there are certain structured data sets, such as when $\X$ is near diagonal, for which LO
is poorly suited. In such cases, {\it generalized cross-validation} (GCV) can sometimes provide better
performance. GCV is closely related to LO; only instead of computing
LO on the observations $(\X, \y)$, GCV first rotates the observations using a complex unitary matrix $\U$,
$$
  \tilde{\X} = \U \X,\qquad \tilde{\y} = \U \y,
$$
where $\U$ is chosen to make 
$\tilde{\X} \tilde{\X}^H$ circulant; then GCV computes LO on the rotated observations $(\tilde{\X}, \tilde{\y})$.
The transformation step makes GCV invariant to rotations of the observations and can fix certain problem cases
with LO
(see [\citegcv]).
As $\tilde{\X}$ and $\tilde{\y}$ are complex matrices, the algorithms developed in this paper are not immediately
applicable, but one might hope that there are versions that could work.

\subend

\medskip
\noindent{\bf References}
\smallskip
\begingroup
\eightpoint
\dimen0=\rightskip
\rightskip = \dimen0 plus 10pt minus 10pt
\item{[\citewishart]} Theodore Anderson. An introduction to multivariate statistical analysis,
third edition. {\sl Wiley-Interscience,} 2003.
\smallskip
\item{[\citecvsurvey]} Sylvain Arlot and Alain Celisse. A survey of cross-validation procedures
for model selection. {\sl Statistical Surveys,} 4:40--79, 2010.
\smallskip
\item{[\citeloci]} Pierre Bayle, Alexandre Bayle, Lucas Janson, and Lester Mackey.
Cross-validation confidence intervals for test error.
{\sl Advances in Neural Information Processing Systems,} 33:16339--16350, 2020.
\smallskip
\item{[\citeriboflavin]} Peter B\"{u}hlmann, Karkus Kalisch, and Lukas Meier. High-dimensional
statistics with a view towards applications in biology. {\sl Annual Review of Statistics
and its Application,} 1:255--278, 2014.
\smallskip
\item{[\citecv]} Prabir Burman. A comparative study of ordinary cross-validation, v-fold
cross-validation and the repeated learning-testing methods.
{\sl Biometrika,} 76(3):503--514, 1989.
\smallskip
\item{[\citelars]} Bradley Efron, Trevor Hastie, Iain Johnstone, and Robert Tibshirani.
Least angle regression. {\sl Annals of Statistics,} 32(2):407--499, 2004.
\smallskip
\item{[\citegeo]} Santosa Fadil and Symes W. William. Linear inversion of 
band-limited reflection seismograms.
{\sl SIAM Journal on Scientific and Statistical Computing,} 7(4):1307--1330, 1986.
\smallskip
\item{[\citegcv]} Gene Golub, Michael Heath, and Grace Wahba. Generalized cross-validation as a method for
choosing a good ridge parameter. {\sl Technometrics,} 21(2):215--223, 1979.
\smallskip
\item{[\citematrix]} Gene H. Golub and Charles F. Van Loan. Matrix computations, fourth
edition. {\sl The John Hopkins University Press,} 2013.
\smallskip
\item{[\citeriboflavinfirst]} Jian-Ming Lee, Shehui Zhang, Soumitra Saha, Sonia Anna,
Can Jiang, and John Perkins. RNA expression analysis using an antisense {\it Bacillus subtilis}
genome array. {\sl Journal of Bacteriology,} 183(24):7371--7380, 2001.
\smallskip
\item{[\citelarslike]} Michael Osborne, Brett Presnell, and Berwin Turlach. On the lasso and its
dual. {\sl Journal of Computational and Graphical Statistics,} 9(2):319--337, 2000.
\item{[\citealo]} Kamiar R. Rad and Arian Maleki. A scalable estimate of the out-of-sample prediction
error via approximate leave-one-out cross-validation. {\sl Journal of the Royal Statistical Society Series B,}
 82(4):965--996, 2020.
\smallskip
\item{[\citelop]} Kamiar R. Rad, Wenda Zhou, and Arian Maleki. Error bounds in
estimating the out-of-sample prediction error using leave-one-out cross-validation
in high-dimensions. {\sl International Conference on Artificial Intelligence and Statistics,} PMLR 2424--2434, 2020.
\smallskip
\item{[\citerobotsrc]} Carl Rasmussen and Christopher Williams. Gaussian processes for machine
learning. {\sl The MIT Press,} 2006.
\smallskip
\item{[\citeconvex]} Ralph Rockafellar. Convex analysis. {\sl Princeton University Press,} 1970.
\smallskip
\item{[\citesparse]} Saharon Rosset, Ji Zhu, and Trevor Hastie. Boosting as a regularized
path to a maximum margin classifier. {\sl Journal of Machine Learning Research,} 5:941--973, 2004.
\smallskip
\item{[\citeseismology]} Seth Stein and Michael Wysession. An introduction to seismology, earthquakes, and 
earth structure. {\sl Blackwell Publishing,} 2003.
\smallskip
\item{[\citelasso]} Robert Tibshirani. Regression shrinkage and selection via the lasso.
  {\sl Journal of the Royal Statistical Society B,} 58(1):267--288, 1996.
\smallskip
\item{[\citeunique]} Ryan J. Tibshirani. The lasso problem and uniqueness.
{\sl Electronic Journal of Statistics,} 7:1456--1490, 2013.
\smallskip
\item{[\citegenerallasso]} Ryan Tibshirani and Jonathan Taylor. The solution path of the generalized lasso.
{\sl Annals of Statistics,} 39(3):1335--1371, 2011.
\smallskip
\item{[\citerobotarm]} Sethu Vijayakumar and Stefan Schaal. Locally weighted projection regression: an
O(n) algorithm for incremental real time learning in high dimensional space.
{\sl Proc.\ ICML 2000,} 1079--1086, 2000.
\smallskip
\item{[\citealop]} Shuaiwen Wang, Wenda Zhou, Haihao Lu, Arian Maleki, and Vahab Mirrokni.
Approximate leave-one-out for fast parameter tuning in high dimensions. {\sl Proceedings of 
the 35th International Conference on Machine Learning,} PMLR 80:5228--5237, 2018.
\smallskip
\item{[\citelo]} Haolin Zou, Arnab Auddy, Kamiar R. Rad, and Arian Maleki. Theoretical
analysis of leave-one-out cross-validation for non-differentiable penalties under
high-dimensional settings. {\sl arXiv preprint arXiv:2402.08543,} 2024.

\endgroup

\bigskip
\ifx\mIncludeGuard\undfined
\input mac/m-impl
\def\subend{\end}
\else
\def\subend{}
\fi

\noindent{\bf A\enspace Benchmarks}
\smallskip
\noindent This appendix provides source code to run the benchmarks referenced in \S2. The benchmarks
use the python package {\tt bbai} version 1.13.0 ({\tt https://github.com/rnburn/bbai})
to run the LO-LARS algorithm, and all benchmarks share the common timing code below.
\smallskip
{\eightpoint\tt\catcode`\_=12
\halign{\hskip\leftskip\indent#\hfil\cr
from bbai.glm import Lasso\cr
import time\cr
import numpy as np\cr
\cr
def bench(X, y, early_exit=np.inf):\cr
\ \ model = Lasso(fit_intercept=True, loo_mode='t',\cr
\ \ \ \ \ \ \ \ \ \ \ \ \ \ early_exit_threshold=early_exit)\cr
\cr    
\ \ \# ignore the first run\cr
\ \ model.fit(X, y)\cr
\cr
\ \ \# average time to fit over 5 runs\cr
\ \ elapse = 0 \cr
\ \ N = 5 \cr
\ \ for \_ in range(N):\cr
\ \ \ \ t1 = time.time()\cr
\ \ \ \ model.fit(X, y)\cr
\ \ \ \ elapse += time.time() - t1\cr
\ \ elapse /= N\cr
\cr
\ \ print('avg elapse:', elapse)\cr
}
}
\medskip
\noindent {\it Diabetes.} The diabetes data set measures diabetes progression in 442 patients over the course
of a year. It comes from [\citelars] and is available from scikit-learn.
\smallskip
{\eightpoint\tt\catcode`\_=12
\halign{\hskip\leftskip\indent#\hfil\cr
from sklearn.datasets import load_diabetes\cr
from sklearn.preprocessing import StandardScaler\cr
import numpy as np\cr
\cr
data = load_diabetes()\cr
X = data['data']\cr
y = data['target']\cr
n, p = X.shape\cr
\cr
\# standard benchmark\cr
bench(StandardScaler().fit_transform(X), y)\cr
\cr
\# quadratic benchmark\cr
pp = p * (p-1)//2 - 1 + 2*p\cr
Xp = np.zeros((n, pp))\cr
cnt = 0\cr
for j, f in enumerate(data['feature_names']):\cr
\ \ Xp[:, cnt] = X[:, j]\cr
\ \ cnt += 1\cr
\ \ \# don't include the squared term for sex\cr
\ \ if f != 'sex':\cr
\ \ \ \ Xp[:, cnt] = X[:, j]**2\cr
\ \ \ \ cnt += 1\cr
\ \ for jp in range(j+1, p):\cr
\ \ \ \ Xp[:, cnt] = X[:, j] * X[:, jp]\cr
\ \ \ \ cnt += 1\cr
bench(StandardScaler().fit_transform(X), y)\cr
}
}
\medskip
\noindent {\it California housing.} This data set provides median house value for different
districts in California. It is available from scikit-learn.
\smallskip
{\eightpoint\tt\catcode`\_=12
\halign{\hskip\leftskip\indent#\hfil\cr
from sklearn.datasets import fetch_california_housing\cr
from sklearn.preprocessing import StandardScaler\cr
from common import bench\cr
import numpy as np\cr
\cr
X, y = fetch_california_housing(return_X_y=True)\cr
n, p = X.shape\cr
\cr
\# standard benchmark\cr
bench(StandardScaler().fit_transform(X), y)\cr
\cr
\# quadratic benchmark\cr
pp = p*(p-1)//2 + 2*p\cr
cnt = 0\cr
Xp = np.zeros((n, pp))\cr
for j in range(p):\cr
\ \ Xp[:, cnt] = X[:, j]\cr
\ \ cnt += 1\cr
\ \ for jp in range(j, p):\cr
\ \ \ \ Xp[:, cnt] = X[:, j] * X[:, jp]\cr
\ \ \ \ cnt += 1\cr
X = Xp\cr
bench(StandardScaler().fit_transform(X), y)\cr
}
}
\medskip 
\noindent {\it Riboflavin.} This data set measures riboflavin production in bacteria cultures. It
original comes from [\citeriboflavinfirst] and was made available by [\citeriboflavin]. The data
can be downloaded as a CSV file from
\hbox{\tt\sevenpoint http://www.annualreviews.org/doi/suppl/10.1146/annurev-statistics-022513-115545}.
\smallskip
{\eightpoint\tt\catcode`\_=12
\halign{\hskip\leftskip\indent#\hfil\cr
import pandas as pd\cr
import numpy as np\cr
from common import bench\cr
\cr
df = pd.read_csv('riboflavin.csv')\cr
y = df.iloc[0, 1:].to_numpy()\cr
X = df.iloc[1:, 1:].to_numpy().T\cr
\# Note: following [\citeriboflavin], the data is assumed to\cr
\# \ \ \ \ \ \ \ already be on the same scale\cr
\cr
\# standard benchmark\cr
bench(X, y)\cr
\cr
\# early exit benchmark\cr
bench(X, y, early_exit=0.01)\cr
}
}
\medskip 
\noindent {\it SARCOS Inverse.} This data set relates to an inverse dynamics problem for a robot arm.
The data set originally comes from [\citerobotarm] and is made available by [\citerobotsrc] as a MAT-file at
\hbox{\tt\sevenpoint https://gaussianprocess.org/gpml/data/}.
\smallskip
{\eightpoint\tt\catcode`\_=12
\halign{\hskip\leftskip\indent#\hfil\cr
import scipy.io\cr
import numpy as np\cr
from sklearn.preprocessing import StandardScaler\cr
from common import bench\cr
\cr
mat = scipy.io.loadmat('sarcos_inv.mat')\cr
X = mat['sarcos_inv']\cr
y = X[:, 21]\cr
X = X[:, :21]\cr
n, p = X.shape\cr
\cr
\# standard benchmark\cr
bench(StandardScaler().fit_transform(X), y)\cr
\cr
\# quadratic benchmark\cr
pp = p*(p-1)//2 + 2*p\cr
cnt = 0\cr
Xp = np.zeros((n, pp))\cr
for j in range(p):\cr
\ \ Xp[:, cnt] = X[:, j]\cr
\ \ cnt += 1\cr
\ \ for jp in range(j, p):\cr
\ \ \ \ Xp[:, cnt] = X[:, j] * X[:, jp]\cr
\ \ \ \ cnt += 1\cr
X = Xp\cr
bench(StandardScaler().fit_transform(X), y)\cr
\cr
\# quadratic, early exit benchmark\cr
bench(StandardScaler().fit_transform(X), y, early_exit=0.01)\cr
}
}

\subend

\bye